%% file: main.tex
\documentclass{article} 
\usepackage{iclr2021_conference,times}

\usepackage{soul}
\usepackage{hyperref}
\usepackage{url}
\usepackage{bbm}
\usepackage{mathtools}
\usepackage{booktabs}  
\usepackage{siunitx}
\usepackage{caption}
\usepackage{subcaption}
\usepackage[colorinlistoftodos,prependcaption,textsize=tiny]{todonotes}
\usepackage{fixltx2e}
\usepackage{wrapfig}
\usepackage{cleveref}
\crefformat{section}{\S#2#1#3} 
\crefformat{subsection}{\S#2#1#3}
\crefformat{subsubsection}{\S#2#1#3}

\input{math_commands}

\input{our-macro}

\usepackage{amsthm}
\newtheorem{theorem}{Theorem}[section]

\theoremstyle{definition}
\newtheorem{definition}{Definition}[section]

\usepackage{enumitem}
\setlist{leftmargin=4mm}

\title{Resurrecting Submodularity for Neural Text Generation}

\author{Simeng Han\thanks{\ \ Equal contribution.} $^\P$, Xiang Lin\footnotemark[1] $^\P$ and Shafiq Joty$ ^\P$$^\S$  \\
  $^\P$Nanyang Technological University, Singapore \\
  $^\S$Salesforce Research Asia, Singapore \\
  \texttt{\{hans0035@e., linx0057@e., srjoty@\}ntu.edu.sg } \\ }

\iclrfinalcopy 
\begin{document}

\maketitle

\begin{abstract}
Submodularity is desirable for a variety of objectives in content selection where the current neural encoder-decoder framework is inadequate. However, it has so far not been explored in the neural encoder-decoder system for text generation. {In this work,} we define diminishing attentions with submodular functions and in turn, prove the submodularity of the effective neural coverage. 
The greedy algorithm approximating the solution to the submodular maximization problem is not suited to attention score optimization in auto-regressive generation. Therefore instead of following how submodular function has been widely used, we propose a simplified yet {principled} solution. The resulting attention module offers an architecturally simple  and empirically effective method to improve the coverage of neural text generation. We run experiments on three {directed} text generation tasks with different levels of recovering rate, across two modalities, three {different} neural model architectures and two training strategy variations. The results and {analyses} demonstrate that our method generalizes well across these settings, produces texts of good quality and outperforms state-of-the-art baselines. 
\end{abstract}


\section{Introduction} \label{sec:intro}
\input{introduction.tex}

\section{Background}\label{sec:background}
\input{background.tex}

\section{Method}\label{sec:method}

\input{method.tex}

\section{Experiments} \label{sec:experiments}
\input{experiments-sum.tex}
\input{experiments-imgpara.tex}

\input{experiments-nmt.tex}

\section{Analysis} \label{sec:analysis}
\input{analysis-attn.tex}
\section{Conclusion} \label{sec:conclusion}
We have defined a class of \emph{diminishing attentions} and in turn, proved the submodularity of the effective neural coverage. {Submodularity is desirable for coverage. To address the problem that greedy selection cannot be utilized over attention scores in the neural framework, we propose a simplfied solution.} Experimental results and a series of analyses on three tasks across two modalities, five datasets and three neural architectures demonstrate that our method produces text outputs of good quality, outperforms comparable baselines and achieves state-of-the-art performance.

\bibliography{iclr2021_conference}
\bibliographystyle{iclr2021_conference}



\newpage
\appendix

\input{appendix.tex}

\end{document}

%% file: math_commands.tex
\usepackage{amsmath,amsfonts,bm}

\def\eqref#1{equation~\ref{#1}}

\def\1{\bm{1}}

\def\vtheta{{\bm{\theta}}}

\DeclareMathAlphabet{\mathsfit}{\encodingdefault}{\sfdefault}{m}{sl}
\SetMathAlphabet{\mathsfit}{bold}{\encodingdefault}{\sfdefault}{bx}{n}

\def\gH{{\mathcal{H}}}

\def\sA{{\mathbb{A}}}

\def\sR{{\mathbb{R}}}
\def\sS{{\mathbb{S}}}
\def\sT{{\mathbb{T}}}

\def\sV{{\mathbb{V}}}

\DeclareMathOperator*{\argmax}{arg\,max}

%% file: our-macro.tex
\usepackage{xspace}

\newcommand{\ie}{{\em i.e.,}\xspace}
\newcommand{\eg}{{\em e.g.,}\xspace}


%% file: introduction.tex
Monotone nondecreasing submodular objectives have been shown to be ideal for content selection and alignment in \emph{extractive} text summarization and \emph{statistical} machine translation, respectively \citep{lin-bilmes-2010-multi,lin-bilmes-2011-class, lin-bilmes-2011-word}. 
Indeed, it can be shown that many popular extractive summarization methods \citep{Carbonell:1998:MMD,berg-kirkpatrick-etal-2011-jointly} optimize a submodular objective. 
Despite their appropriateness, submodular functions for content selection have so far been ignored in neural text generation models.

The neural encoder-decoder framework trained in an end-to-end manner maintains the state-of-the-art (SoTA) in a class of {directed} text generation tasks aimed at recovering the source message either to the full or a compressed version of it. A major shortcoming of such architectures in dealing with text generation is that they could keep covering some parts in the source while ignoring the other important concepts, thus resulting in less comprehensive coverage. Various mechansims for improving the neural coverage have been shown to be effective \citep{see-etal-2017-get,tu-etal-2016-modeling,google-nmt}. However, they either require extra parameters and loss to furnish the model with better learning capacity or place a specific bound on the sum of attention scores. 

In this work, we define a class of novel attention mechanisms called \textit{diminishing attentions} with submodular functions and in turn, prove the submodularity of the effective neural coverage. The submodular maximization problem is generally approximated by greedy selection. However, it is not suited to optimizing attention scores in auto-regressive {generation} systems. We therefore put forward a simplified yet {principled and empirically effective} solution. By imposing \emph{submodularity} on the coverage enforced by the decoder states on the encoder states, our {diminishing attention} method enhances the model's awareness of previous steps, leading to more comprehensive overall coverage of the source and maintaining a focus on the most important content when the goal is to generate a compressed version of the source (\eg\ text summarization). We further enhance our basic {diminishing attention} and propose \textit{dynamic diminishing attention} to enable dynamically adapted coverage. 
Our results highlight the benefits of submodular coverage. Our diminishing attention mechanisms achieve SoTA results on three diverse {directed} text generation tasks, {abstractive summarization, neural machine translation (NMT)} and image-paragraph generation spanning across two modalities, three neural architectures and two training strategy variations.



%% file: background.tex
\subsection{Submodular functions} \label{subsec:submodular}

Let $\sV=\left\{v_{1}, \ldots, v_{n}\right\}$ denote a set of $n$ objects and $f: 2^{\sV} \rightarrow \sR$ is a  set-function that returns a real value for any subset $\sS \subseteq \sV$. We also assume $f(\phi) = 0$. Our goal is to find the subset:
\begin{equation}
    \sS^* = \argmax_{\sS \subseteq \sV} f(\sS) \hspace{2em} \text{s.t. } |\sS^*| \leq m \label{eq:comb}
\end{equation}
\noindent where $m$ is the budget; \eg\ for summarization, $m$ is the maximum summary length allowed. Note that $f: 2^{\sV} \rightarrow \mathbb{R}$ can also be expressed as $f: \{0,1\}^n \rightarrow \mathbb{R}$, where a subset $\sS \subseteq \sV$ is represented as a one-hot vector of length $n$, that is, $\sS = (\mathbbm{1}(v_1 \in \sS), \ldots, \mathbbm{1}(v_n \in \sS))$ with $\mathbbm{1}$ being the indicator function that returns $1$ if the argument is true otherwise $0$. In general, solving Equation \ref{eq:comb} is NP-hard. Even when $f$ is \emph{monotone submodular} (defined below), it is still NP-complete. 

\theoremstyle{definition}
\begin{definition}\label{def:1} 
$f$ is submodular if $f(\sS + v) - f(\sS) \ge f(\sT + v) - f(\sT)$ for all $\sS \subseteq \sT \subseteq \sV, v \notin \sS$. 
\end{definition}
\vspace{-0.3em}
\noindent This property is also known as \emph{diminishing returns}, which says that the information gain given by a candidate object (\eg\ a word or sentence) is larger when there are fewer objects already selected (as summary). The function $f$ is \emph{monotone nondecreasing} if for all $\sS \subseteq \sT$, $f(\sS) \le f(\sT)$. In this paper, we will simply refer to monotone nondecreasing submodular functions as submodular functions. 



Submodular functions can be considered as the discrete analogue of
concave functions in that $f(\vtheta): \sR^n \rightarrow \sR$ is concave if the derivative $f'(\vtheta)$ is non-increasing in $\vtheta$, and $f(\sS): \{0,1\}^n \rightarrow \mathbb{R}$ is submodular if for all $i$ the discrete derivative,  $\partial_i f(\sS) = f(\sS + v_i) - f(\sS)$ is non-increasing in $\sS$. Furthermore, if $g: \sR_{+} \rightarrow \sR$ is concave, then the composition $f^{'}(\sS) = g (f(\sS)): 2^\sV \rightarrow \sR$ is also submodular. The convex combination of two submodular functions is also submodular. 

\subsection{Neural coverage}
Neural coverage of one encoder state can be defined as the sum of the attention scores that it receives over the first until the previous decoding step \citep{google-nmt, tu-etal-2016-modeling}. Formally, the coverage of encoder state $i$ at decoding step $t$ is $c_{i}^{t}=\sum_{t^{\prime}=0}^{t-1} a_{i}^{t^{\prime}}$, where $a_{i}^{t^{\prime}}$ are the attention scores. In abstractive summarization, \citet{see-etal-2017-get} use coverage to keep track of what has been generated so far by assigning trainable parameters to the coverage and using it to guide the attention module in making decisions. They also introduce a coverage loss to discourage the network from repeatedly attending to the same parts, thereby avoiding repetition in the generated summary. In NMT, \citet{google-nmt} apply a coverage penalty during decoding which restricts the coverage of an input token from exceeding 1. \citet{tu-etal-2016-modeling} maintain a coverage vector which is updated with the recurrence unit and fed into the attention model. The major differences between our method and the previous methods are that we do not require extra parameters or extra losses to furnish the network with better learning capacity, and we do not place a specific bound on the sum of attention scores. Moreover, the effective neural coverage of our method is submodular. 


%% file: method.tex
In this section, we present \textit{submodular coverage} and our  \textit{diminishing attentions} for the neural encoder-decoder model, and {we show the \textit{effective coverage} based on the diminishing attentions.} 

\subsection{Submodular coverage} \label{subsec:subcov}
In the general encoder-decoder framework, the input is represented as a set of latent states (concepts) from an encoder, and the decoder constructs the output autoregressively by generating one token at a time. While generating a token, the decoder computes an {attention} distribution over the encoded latent states, which represents the relevance of the corresponding input to the output token. 

Following previous work \citep{google-nmt,see-etal-2017-get}, we quantify the degree of {coverage} of an encoder state as the sum of the set of attentions that the decoder puts on {the state} in the course of generating the output sequence.
Let us consider adding a new token $w$ into two outputs $\mathbb{S}$ and $\mathbb{S}^{\prime}$, where the concepts covered by $\mathbb{S^{\prime}}$ is a subset of those covered by $\mathbb{S}$. Intuitively, the information gain from adding $w$ to $\mathbb{S}^{\prime}$ should be higher than adding it to $\mathbb{S}$, as the new concepts carried by $w$ might have already been
covered by those that are in $\mathbb{S}$ but not in $\mathbb{S}^{\prime}$. This is indeed the \emph{diminishing return} property. 

We thus put forward our hypothesis on a desirable property of the neural coverage function that it should be submodular. The greedy algorithm proposed by \citet{Nemhauser:1978} approximates the solution to the submodular maximization problem (Eq. \ref{eq:comb}) with an optimality of $0.63$ or higher. For that purpose, the attention scores should be added to the coverage in a greedy manner. However, greedy search among all the  states is not possible when the decoder states are generated autoregressively, one at a time.
We therefore propose a simplified and principled solution as detailed below.

Let $\sA_{i} = \left\{a_{i}^{0}, \dots, a_{i}^{t}\right\}$ denote the set of attention scores that an encoder state $i$ receives from the first ($t=0$) till the current decoding step $t$, and 
$F: 2^{\sA_{i}} \rightarrow \mathbb{R}$ be a set function that maps these scores to a score which we define as \textit{submodular coverage} at the current step $t$.\footnote{For convenience of developing our method, we define $c_i^{t}$ as the sum of attentions from the first until the current decoding step instead of the previous decoding step.} 
\theoremstyle{definition}
\begin{definition}
\textbf{Submodular coverage:} 
\vspace{-0.6em}
\begin{equation}
    F(\sA_{i}^{t}) = g(\sum_{t^{\prime}=0}^{t} a_{i}^{t^{\prime}}) + b
    \label{eq:subcov}
\end{equation}
\end{definition}
\vspace{-0.6em}
\noindent where $g$ is a concave and non-decreasing function (\eg\ $\log(x+1)$, $\sqrt{x+1}$), and $b$ is a constant and equal to $-g(0)$. $F$ is monotone submodular because it imposes a concave function on the modular or additive coverage function $f=\sum_{t^{\prime}=0}^{t} a_{i}^{t^{\prime}}$ (see the composition property mentioned in  \Cref{subsec:submodular}).

\begin{figure*}[t!]
    \centering
    \includegraphics[width=1\textwidth]{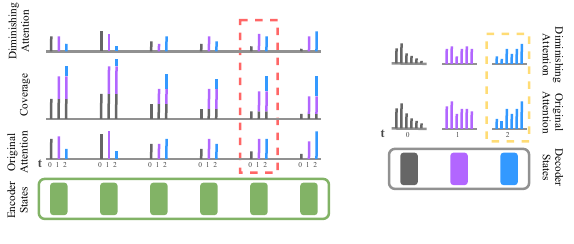}
    \vspace{-0.9em}
    \caption{Illustration of diminishing attention in an encoder-decoder model over the encoder states. The brown, purple and blue bars are for the first, second and third decoding step respectively. The left figure shows coverage, original attention and diminishing attention for decoding steps $t=0$ to $2$. 
    The \textbf{red dashed block} shows how diminishing attention is related to original attention and coverage for the same single encoder state. For example, at decoding step $2$, even though original attention has the same value as that at step $1$, the diminishing attention is smaller since the encoder state has been covered more from steps $0$ to $1$. 
    The right figure shows a summary of the original and diminishing attentions on the encoder states of the left figure in the first, second and third decoding steps. 
    The \textbf{yellow dashed block} shows
    at decoding step $2$, the effective attentions for encoder states with higher coverage (\eg \ the first encoder state) have diminished more than those with lower coverage (\eg \ the last encoder state). }
    \label{fig:model}
\vspace{-2.0em}
\end{figure*}

\subsection{Diminishing attention}
By subtracting the submodular coverage between the current step and the previous step, we model diminishing attention scores based on the original attention. 
Formally, \textit{diminishing attention} (DimAttn) is defined as 
\vspace{-0.2em}
\begin{equation}
    \textbf{DimAttn}_{i}^{t} = F(\mathbb{A}_{i}^{t}) - F(\mathbb{A}_{i}^{t-1}),
    \label{eq:sub_attn}
\end{equation}
\noindent which models \emph{diminishing return} directly, and will be used as the attention weight corresponding to the encoder state ${i}$ to produce the context vector at decoding step $t$ to predict the next token.

Thus the effective attention scores are optimized with a submodular function. The diminishing return property of $F$ in Eq. \ref{eq:sub_attn} realizes the effect that if an encoder state $i$ receives the same amount of attention at two different decoding steps $t$ and $t^{\prime}$ such that $t^{\prime} > t$, the effective attention would diminish more at $t^{\prime}$ because the coverage at $t^{\prime}$ is larger. Furthermore, because $
g$ is concave, when two {encoder states $i$ and $j$} have different amounts of coverage at step $t-1$, and they receive the same attention score at step $t$, the {state} with a larger coverage from previous steps would receive a smaller effective attention. We visualize these two properties of diminishing attention in Figure~\ref{fig:model}.

\paragraph{Effective coverage.}  The \textit{effective coverage} of an encoder state is the sum of effective attention scores that it receives from the first till the current decoding step.

\begin{theorem}
The {effective coverage} with diminishing attention is submodular. 
\end{theorem}
\vspace{-1em}
\begin{proof}
{Let the effective coverage of an encoder state $i$ at decoding step $t$ be ${ec}_{i}^{t}$, then we can show}
\vspace{-0.5em}
\begin{equation}
\begin{split}
        {ec}_{i}^{t} = \sum_{t^{\prime}=0}^{t}\text{DimAttn}_{i}^{t^{\prime}} 
        &= \sum_{t^{\prime}=0}^{t}(F(\mathbb{A}_{i}^{t^{'}}) - F(\mathbb{A}_{i}^{t^{'}-1})) 
        = F(\mathbb{A}_{i}^{t}) - F(\emptyset{}) = F(\mathbb{A}_{i}^{t}) 
\end{split}
\label{eq:derivation}
\end{equation}
where all the terms in between get cancelled. 
Since $F(\mathbb{A}_{i}^{t})$ is submodular, ${ec}_{i}^{t}$ is also submodular.  
\end{proof}
\vspace{-1em}

To emphasize, the effective coverage that each encoder state {acquires} from the decoder at every decoding step is equal to the submodular coverage defined in Eq. \ref{eq:subcov}, while coverage is apparently modular with attention. 
Additionally, since $g$ is monotone non-decreasing, it is guaranteed that although the coverage has been changed, the encoder states which receive the largest coverage with the original attention still receive the largest effective coverage with the diminishing attention.

\subsection{Dynamic diminishing attention}


Using a single submodular coverage function alone may not yield the most appropriate diminishing return effect of the coverage for each encoder state in the decoding process {due to the lack of flexibility}. More ideally, the model should be capable of further adopting varied degrees of diminishing effect as the decoding proceeds. 

Let $F_1\left(\mathbb{A}_{i}^{t}\right)=g_1\left(\sum_{t^{\prime}=0}^{t} a_{i}^{t^{\prime}}\right)+b_1$ and $F_2\left(\mathbb{A}_{i}^{t}\right)=g_2\left(\sum_{t^{\prime}=0}^{t} a_{i}^{t^{\prime}}\right)+b_2$  be two different submodular coverage functions. 
We assume $g_1$ has a smaller first-order derivative than $g_2$,
thus given the same $A_i^t$, the diminishing effect of the submodular coverage $F_1$ would be stronger than that of $F_2$. 


If an encoder state has received a particularly large attention at a certain step, the weight of the more aggressive diminishing function should increase. We thus compute the probability of applying a more aggressive diminishing function at step $t$ as $P_i^t = \text{max}_t(\mathbb{A}_i^{t-1})$. 
We use it to dynamically control the relative weights of the diminishing functions. 
The \textbf{dynamic diminishing attention} (DyDimAttn) is thus defined as:
\begin{equation}
        \textbf{DyDimAttn}^t_i = P_i^t [F_1(\mathbb{A}_i^t) - F_1(\mathbb{A}_i^{t-1})]\\
    + (1 - P_i^t)[F_2(\mathbb{A}_i^t) - F_2(\mathbb{A}_i^{t-1})]
    \label{eq:Dydimattn}
\end{equation}
\noindent
which is a convex combination of two diminishing attentions, where the diminishing attention which diminishes faster is weighted with $P_i^t$ and the other weighted with $(1 - P_i^t)$. 

Since $P_i^t$ keeps changing, the proof of effective coverage of diminishing attention (Eq. \ref{eq:derivation}) is not suited for dynamic diminishing attention. Thus we prove the submodularity of the effective coverage of dynamic diminishing attention with the definition of submodular functions.
\begin{theorem}
The \textbf{effective coverage} of dynamic diminishing attention is submodular. 
\end{theorem}

\vspace{-1em}
\begin{proof}
{The effective coverage of an encoder state $i$ at  step $t$ with dynamic diminishing attention is}
\vspace{-0.5em}
\begin{equation}
     ec_i^t = \sum_{t^{\prime}=0}^{t} \text{DyDimAttn}_{i}^{t^{\prime}}
\end{equation}
Coverage increases as the set $A_i^{t}$ gets larger over steps $1$ to $t$ and it is obvious that $P_i^t$ is monotone non-decreasing over steps $1$ to $t$. 
The return of adding the same amount {of} original attention score to the set $A_i^{t}$ is smaller at a later step as the weight  over the concave function which diminishes faster ({$P_i^t$}) becomes larger over steps $1$ to $t$.
Thus by definition \ref{def:1}, the effective coverage of dynamic diminishing attention is submodular. 
\end{proof}
\vspace{-1.5em}

%% file: experiments-sum.tex
From the perspective of coverage, text summarization aims to recover a compressed version of the source document, concentrating on the most important concepts; image-paragraph generation aims to recover descriptions of the image regions while ignoring minor details; and MT aims to recover the full of the source, articulating every detail. In this section, we show that diminishing attentions improve the performance of these three tasks with different levels of recovering rate. 
{Other than the method-specific ones, we use the same hyper-parameters as the baseline for most settings. See \Cref{append:implementation} for more implementation details.} 

\subsection{Abstractive text summarization} \label{subsec:pgexp}
Abstractive summarization involves generating novel phrases to cover the most important information of the input document in a human-like fashion. State-of-the-art pretraining-based abstractive summarization models \citep{lewis2019bart, yan2020prophetnet} suffer from the problem of having repetitive phrases in the output, which has been addressed by blocking duplicated trigrams {during inference} \citep{Paulus2018ICLR}. 

\vspace{-0.5em}
\paragraph{Setup.}
We use two benchmark news summarization datasets following standard splits: CNN/DM \citep{hermann2015CNN,nallapati-etal-2016-abstractive} and NYT50 \citep{durrett-etal-2016-learning}.

\textit{CNN/DM.} On CNN/DM, we first evaluate our method based on the LSTM based Pointer-Generator (PG) model \citep{see-etal-2017-get} which we fine-tune with our diminishing attentions.   
Following the original setting, the source article is truncated to 400 tokens in training the baseline PG models. Inclusion of more input tokens does not give additional gain to the baselines, whereas exposing the models to more input tokens was beneficial {on the validation set} when diminishing attentions were employed. We truncate the source article to 600 tokens for training with {DimAttn} and 800 tokens for {DyDimAttn}. We include a comparison in the same 400-token setup in Appendix \ref{append:400-ablation}. 

{On CNN/DM, we also evaluate our attentions within the recently proposed SoTA model BART \citep{lewis2019bart}, where we replace the last 7 layers of encoder-decoder cross attention with our diminishing attentions and finetune it.}
Given the same set of hyper-parameters, it takes around 3.81 seconds for training each batch with the original attention and around 3.97 seconds for training each batch with the diminishing attention on 1 RTX2080 GPU. This means that training the diminishing attention takes only around $0.42\%$ extra amount of time for training one batch. 

\textit{NYT50.} On NYT50, we evaluate based on the SoTA BERT-based Transformer model \citep{Liu2019TextSW} by replacing the encoder-decoder cross attention at the last layer of the decoder with our attentions and fine-tuning it. 


 We evaluate the performance with F1 ROUGE \citep{rouge} and \textbf{MoverScore}  \citep{zhao-etal-2019-moverscore}, which {is} Earth Mover distance based on {BERT \citep{devlin-etal-2019-bert}} contextual embeddings. 

\vspace{0.1em}
\paragraph{Results.}
\textit{ROUGE scores.} 
We show the results of PG-based models and BART-based models on CNNDM in Table~\ref{table:cnn/dm}. Our method is effective on both LSTM and BART-based models, showing its generalizability across network architectures. In the third block, we show results of recent state-of-the-art models on CNNDM and our model {based on BARTSum}  outperforms all of them.

\noindent\textit{MoverScore.} 
In Tables \ref{table:cnn/dm} and \ref{tab:nyt}, we have also shown {MoverScore} results for the models on the respective datasets. The consistent improvements in MoverScore show that models equipped with diminishing attentions are capable of generating outputs more semantically similar to the gold summary than the baselines. This indicates that our method is more effective in capturing the overall meaning of the source article than the baselines.

\begin{table}[t]
\centering 
\caption{ROUGE F1 score and {MoverScore} (1-gram and 2-grams) results on CNN/DM. We also report the size of pretraining data (P-Data) and parameters (Params) of each model. }
\resizebox{1\columnwidth}{!}{%
\begin{tabular}{lccccccc}
\toprule
        & \multicolumn{3}{c}{\textbf{ROUGE Scores}} 
        & \multicolumn{2}{c}{\textbf{MoverScore}} \\
 \cmidrule(lr){2-4}   \cmidrule(lr){5-6}  
 \cmidrule(lr){7-8}    
 \textbf{Model} &\bf{R-1} &\bf{R-2} &\bf{R-L} & \textbf{1-gr} & \textbf{2-gr}  & \textbf{P-Data} & \textbf{Params} \\
\midrule
LEAD-3                         & 40.00 & 17.50 & 36.28 & -- & --  \\
\midrule
\textbf{LSTM-based} \\ 
PG                    & 36.69  & 15.92 &  33.63  & 12.46 & 19.37  & 0 & 27M \\
PG + Cov.             & 39.08 & 17.09 & 35.92   & 17.55 &  24.17 & 0 & 27M + 512 \\
PG + Dim            & 40.01 & 17.74 & 36.94  & 17.56 &  24.16 & 0 & 27M \\
PG + DyDim & \textbf{40.13} & \bf{17.94} & \bf{37.21}  & \bf{17.77} & \bf{24.38} & 0 & 27M \\
\midrule
\textbf{Large-Size Pretrained Models} \\ 
ERNIE-GEN \citep{xiao2020erniegen} & 44.02 & 21.17 & 41.26 &  -- & -- & 16G & 340M\\
PEGASUS\textsubscript{C4} \citep{zhang2019pegasus} & 43.90 & 21.20 & 40.76  & -- & -- & 750G & 568M\\
PEGASUS\textsubscript{HugeNews} \citep{zhang2019pegasus} & 44.17 & 21.47 & 41.11 &  -- & -- & 3.8T & 568M\\
ProphetNet \citep{yan2020prophetnet}  & 44.20 & 21.17 & 41.30 &  -- & -- & 160G & 400M\\ 
BARTSum \citep{lewis2019bart} & 44.16 & 21.28 & 40.90 & 22.34 & 28.47 & 160G & 400M \\
{BARTSum + Dim}    & 44.86 & \bf{21.76} & 41.62 & 23.05 & 29.04 & 160G & 400M \\
{BARTSum + DyDim}  & \bf{44.92} & 21.70 & \bf{41.66} & \bf{23.12} & \bf{29.13} & 160G & 400M \\
\bottomrule
\end{tabular}
}
\label{table:cnn/dm}
\end{table}

\begin{figure}[t!]
\begin{minipage}[t]{0.5\textwidth}
\centering
\captionof{table}{
ROUGE Scores and MoverScores results on the NYT50 summarization dataset. 
}
\label{tab:nyt}
\resizebox{0.97\columnwidth}{!}{
\begin{tabular}{lccccc} 
\toprule
        & \multicolumn{3}{c}{\textbf{ROUGE Scores}} 
        & \multicolumn{2}{c}{\textbf{MoverScore}} \\
         \cmidrule(lr){2-4}   \cmidrule(lr){5-6} 
        \textbf{Model}
        & \textbf{R-1} & \textbf{R-2} & \textbf{R-L}
        & \textbf{1-gr} & \textbf{2-gr} \\
        \midrule 
        LEAD-3                & 24.52 & 12.78 & 21.75 & -- & -- \\
        BertSum                    & 48.33 & 31.03 & 44.85  &  28.16 & 34.09 \\ 
        + Dim                 & 49.29 & \textbf{31.72} & 45.78 & 29.10 & 34.87 \\
        + DyDim               & \textbf{49.46} & 31.59 & \textbf{45.94} & \textbf{29.24} &  \textbf{35.00} \\
        \bottomrule
        \end{tabular}
        }
\end{minipage}
\hfill
\begin{minipage}[t]{0.48\textwidth}
\centering
\captionof{table}{CIDEr results based on two training regimes (Cross-Entropy and Self-Critical) on the Stanford Image-Paragraph dataset.}
\label{tab:imgpara-results}
\resizebox{0.97\columnwidth}{!}{%
\begin{tabular}{lccc}
\toprule
\textbf{Model} &\bf{Cross-Entropy} &\bf{Self-Critical}\\
\midrule
Baseline & 22.68 & 30.63\\
+Dim    & 25.47 & 33.15 \\
+DyDim   & \textbf{25.49} & \textbf{33.28} \\
\bottomrule
\end{tabular}
}
\end{minipage}
\vspace{-1em}
\end{figure}

%% file: experiments-imgpara.tex
\subsection{Image-paragraph generation}
Image-paragraph generation aims to generate a coherent paragraph to describe different aspects of an input image. The widely-used Stanford Image Paragraph dataset \citep{hier-dataset} has been known to be too small in size for the model to learn the language structure and pattern. Previous work \citep{melas-kyriazi-etal-2018-training} has shown that the generated paragraphs usually contain many repetitive phrases and sentences while covering the source poorly. We follow \citep{melas-kyriazi-etal-2018-training} in using the Top-Down model from \citep{top-down} with a CNN pretrained for object detection and a 1-layer LSTM as the language decoder with top-down attention applied over max-pooled features of 40-100 regions of interests (RoI), which we replace with our diminishing attentions. We run on the Stanford Image-Paragraph dataset using the standard splits. We fine-tune on two standard baselines \citep{melas-kyriazi-etal-2018-training}  with different training strategies including minimizing the cross entropy loss and optimizing the CIDEr \citep{CIDEr} reward with self-critical {reinforcement} learning. From the results in \Cref{tab:imgpara-results}, we observe 2.81 improvement in CIDEr \citep{CIDEr} over the cross entropy training baseline, and 2.65 improvement over the self-critical training baseline, setting a new state-of-the-art.

%% file: experiments-nmt.tex
\subsection{Neural machine translation}

We incorporate our attention mechanisms into the cross-attention at the last decoder layer of the Transformer-Big model \citep{scaling_nmt_ott2018scaling} which consists of a 6-layer transformer and fine-tune the pretrained baseline model. 
We run on the WMT'14 English-German (En-De) and WMT'14 English-French (En-Fr) tasks following the settings of \citet{scaling_nmt_ott2018scaling}. As shown in Table~\ref{table:nmt-results}, we obtain 0.4 improvement on En-De and 0.3 improvement on En-Fr in standard tokenized BLEU, which to our knowledge are state of the art without using extra monolingual data  \citep{understanding_backtranslation_scale,Zhu2020Incorporating} or parse tree information \citep{Nguyen2020Tree-Structured}. {We also compute statistical significance for the difference in BLEU scores between our model and the Transformer using paired bootstrap resampling \citep{koehn-2004-statistical}. We conclude that the diminishing attention and dynamic diminishing attention are better than the baseline with at least 99.5\% statistical confidence. }

\begin{wraptable}{r}{6.0cm}
\vspace{-1.3em}
\centering
\caption{BLEU scores and statistical \textbf{conf}idence \citep{koehn-2004-statistical} on  WMT newstest2014 for
English-German and English-French
translation tasks.}
\label{table:nmt-results}
\resizebox{0.41\columnwidth}{!}
{
\begin{tabular}{lcccc} 
\toprule
\textbf{Model} &\bf{En-De \scriptsize (conf)} &\bf{En-Fr \scriptsize (conf)} \\
\midrule
Transformer-Big            & 29.3 & 43.2 \\
\cite{wu2018pay} (reported)   & 29.7 & 43.2 \\ 
Transformer Big +Dim     & 29.7 \scriptsize (99.5) & 43.4 \scriptsize(99.5) \\
Transformer Big +DyDim   & \textbf{29.7} \scriptsize(99.7) & \textbf{43.5} \scriptsize(100.0) \\
\bottomrule
\end{tabular}
}
\vspace{-0.5em}
\end{wraptable}

We conduct further analysis on the WMT En-De task. We first compare the entropy of the normalized effective coverage across all the encoder states at the end of inference, which is denoted as $\gH$ ($t$ is the final step and $\sA_i^t$ contains all the attention scores of encoder state $i$).
\begin{equation*}
    \gH = -\sum_i ec_i^t \log ec_i^t \hspace{2em} \text{ where} \hspace{0.5em}  ec_i^t = ec_i^t/\sum_i ec_i^t
\vspace{-0.3em}
\end{equation*}
and we take the average of $\gH$ of all the test instances. 
From \Cref{table:nmt-analysis}, we see that entropy of the effective coverage of our attentions are higher than that of the baseline. This indicates that the effective coverage distribution of our method is more even across the encoder states than that of the baseline, which suggests that more concepts of the source are covered and the coverage is improved. 

Next, we compare the uni- and bi-gram repetition rates in percentage {computed with the duplicate n-grams in a summary} and see that repetitions become lower with our attentions. Finally, we sort the source sentences in the testset by length and split it into two halves -- \emph{short} and \emph{long}. We observe that our method has more BLEU gains on the longer half. Intuitively, longer source sentences are in more need of even effective coverage to ensure each and every detail of the source is translated. 

\begin{table}[t!]
\caption{Effective coverage entropy $\gH$, repetition rate and BLEU score on subsets with short and long source sentences on English-German newstest2014.\\}
\centering 
\vspace{-1em}
\resizebox{0.8\columnwidth}{!}{\begin{tabular}{lcccccc} 
\toprule
        & \multicolumn{1}{c}{\textbf{Entropy}} 
        & \multicolumn{2}{c}{\textbf{Repetition}}
        & \multicolumn{3}{c}{\bf{BLEU}} \\
 \cmidrule(lr){2-2}   \cmidrule(lr){3-4}  \cmidrule(lr){5-7}  
\textbf{Model} &\bf{$\gH$} & \bf{uni-rep(\%)} & \bf{bi-rep(\%)} & \textbf{short} & \textbf{long} & \textbf{overall} \\
\midrule
Reference & -- & 4.46 & 0.08 & -- & -- & -- \\
Transformer-Big             & 2.08 & 5.87 &  0.18  & 28.9 & 29.5 & 29.3 \\
Transformer-Big + Dim   & 2.37 & 5.82 &  0.16  & 29.1 & 29.8 & 29.7 \\
Transformer-Big + DyDim & 2.41 & 5.85 &  0.16  & 29.1 & 29.9 & 29.7 \\
\bottomrule
\end{tabular}
}
\vspace{-1em}
\label{table:nmt-analysis}
\end{table}

%% file: analysis-attn.tex
\vspace{-0.5em}
We provide more analysis of our method taking summarization on CNN/DM as a case study.

\begin{wraptable}{r}{5.5cm}
\vspace{-1.2em}
\centering
\caption{\textbf{N-gram overlaps} with lead-3.}
\label{table:lead}
\resizebox{0.4\columnwidth}{!}{
\begin{tabular}{lccc} 
\toprule
\textbf{Model} &\bf{R-1} &\bf{R-2} &\bf{R-L}\\
\midrule
Reference         & 40.00 & 17.50 & 36.19 \\
PG + Cov.         & 54.40 & 42.54 & 52.25 \\  
\midrule
PG + Dim          & \textbf{53.02} & \textbf{39.97} & \textbf{50.75} \\
PG + DyDim        & 53.08 &  40.25 & 50.93 \\
\bottomrule
\end{tabular}
}
\vspace{-1.2em}
\end{wraptable}

\subsection{Quantitative and qualitative analysis}
\vspace{-0.5em}
We empirically analyze that submodularity imposed on the coverage enables our models to generate summaries with better coverage of the source document from two aspects: \emph{layout bias} and \emph{repetition ratio}. We also compare our method with trigram-blocking \citep{Paulus2018ICLR}. 

\vspace{-0.5em}
\paragraph{Layout bias.}
\textit{Layout bias} is a common issue in news datasets where the leading section of an article contains the most important information \citep{kryscinski-etal-2019-neural}, and encoder-decoder models are prone to remembering this pattern and ignoring other important content in the rest of the article \citep{kedzie-etal-2018-content}.
Truncating the documents to 400 tokens caters to this bias. 
By increasing the maximum encoding steps
to 600 for DimAttn and 800 for DyDimAttn (\cref{subsec:pgexp}),
we feed the model more information and allow it to automatically learn to extract important information from a longer source.  Table \ref{table:lead} shows that our models have less n-gram overlaps with lead-3 sentences compared to the baseline without compromising the ROUGE scores. This indicates that diminishing attentions enable more comprehensive coverage of the source while maintaining a focus on the most important content. 

\vspace{-1em}
\paragraph{Repetition}
\emph{Trigram-blocking} is widely used for eliminating redundancy in summarization \citep{Liu2019TextSW, gehrmann-etal-2018-bottom}. However, blocking alone does not guarantee high quality as it is not learned. In our analysis of the output from the PG + Cov baseline, we observe that although adopting trigram-blocking results in less repetition and higher ROUGE, the generated summaries are excessively extractive while our method leads to less repetition and more abstractiveness. Our method also improves the abstractiveness of the BERT-based model (see \Cref{append:examples} and \Cref{append:abstractiveness_ratio} for examples and statistics).  

\begin{wraptable}{r}{7cm}
\vspace{-1.5em}
\centering 
\caption{Human evaluation results on Representativeness, Readability and Factual Correctness. Human Agr. is the percentage agreement.}
\label{table:human}
\resizebox{0.5\columnwidth}{!}
{\begin{tabular}{lccc} 
\toprule
\bf{Model}  &\bf{Repr. Win} &\bf{Read. Win} &\bf{Fac.}\\
\midrule
BARTSum                   & 35.9\% & 42.5\% & 96.3\% \\
DyDim       & \textbf{57.2\%} & \textbf{45.1\%} &  \textbf{96.9\%} \\  
Tie                     & 6.87\% & 12.4\% & --\\
\midrule
Human Agr. & 64.1\% & 62.1\% & 90.0\%  \\ 
\midrule 
$p$-value (sign test) & 1e-6 & 0.0014  &  -- \\
\bottomrule
\end{tabular}
}
\vspace{-1.2em}
\end{wraptable}

\subsection{Human evaluation}

\vspace{-0.5em}
We conducted a user study on Amazon Mechanical Turk for the BART-based models. We randomly sampled 500 examples from the CNN/DM test set where each example was distributed to 3 US workers. Each worker was asked to evaluate \emph{representativeness}, \emph{readability} and \emph{factual correctness} of the {system} summaries. We provided the following definition of representativeness and readability as guidelines to the workers: representativeness refers to how well the summary covers the most significant concepts in the source, more specifically, the summary should cover the important concepts and maintain conciseness at the same time; readability is defined as \emph{grammaticality and coherence} where the annotators evaluate the text quality, \ie\ being fluent, logical, consistent and uunderstandable. Annotators were presented with two randomly ordered summaries and asked to pick the better one (win) or equally good (tie) in terms of representativeness and readability and evaluate if the summaries are factually correct (more details about the study are in \Cref{append:human_eval_interface}). We show the human agreement percentage and $p$-value of sign test to assess whether the differences between our model and the baseline were significant. 
From the results in \Cref{table:human}, we notice that our model has significantly better representativeness, reinforcing our hypothesis that the coverage of abstractive summarization should be submodular. 
We also found that our method increases readability and maintains the factual correctness of the baseline. 




\subsection{Discussion}
\vspace{-0.5em}
With regard to neural text generation models aimed at recovering the source, the notion of coverage is more imperative when a model generates repetitive concepts while trying to recover the source.
This could possibly coincide with that the decoder degenerates when it has yet to be sufficiently trained. For example, in image-paragraph generation, the decoder language model may not be well-trained because of the insufficient data. We also notice that models produce highly repetitive outputs at {the early stage of  training}. 
However, diminishing attentions also effectively improve the performance of large-scale models including the BARTSum Transformer model for summarization  with 400M parameters and the Transformer-Big model for machine translation with 220M parameters. This shows that even with the power of transfer learning brought by the {giant pre-trained encoder-decoder model for BARTSum}, or the huge size of parameters and dataset for Transformer-Big, the existing encoder-decoder attention mechanism may not incorporate the most appropriate inductive bias for coverage in these tasks, while our method, by making a simple architectural change to the attention mechanism, effectively improves their performance without adding new parameters. We hypothesize this is because that diminishing attentions explicitly model the submodularity of the neural coverage, {which has been shown a natural fit for content selection \citep{lin-bilmes-2011-class, lin-bilmes-2011-word}.} 

%% file: appendix.tex
\section{Implementation details}
\label{append:implementation}
We use NVIDIA RTX 2080Ti for training PG-based and BERT-based summarization models and the En-De neural machine translation (NMT) models, NVIDIA Tesla V100 for training the BART-based summarization models and En-Fr NMT models, and NVIDIA GTX1080Ti for training image-paragraph generation models. All models are trained with a single GPU. 

\subsection{Abstractive summarization}
\paragraph{Datasets}
We use the entity non-anonymized version of CNN/DM, and use the same data preprocessing on CNN/DM and NYT as the baseline models including the PG-based models{\footnote{\url{https://github.com/abisee/cnn-dailymail}}}, BERT-based models\footnote{\url{https://github.com/nlpyang/PreSumm}} and BART-based models\footnote{\url{https://github.com/pytorch/fairseq/blob/master/examples/bart/README.summarization.md}}. Our train/validation/test split of NYT is the same as {those of} \citep{Liu2019TextSW}. See Table~\ref{tab:dataset_stats} for the dataset statistics.
 
\paragraph{PG-based models}{We train the PG models --- PG and PG with Coverage (PG + Cov.), as our baselines using the settings from \citet{see-etal-2017-get}.} We use $g = g_1 =  \log(x+1)$ for diminishing attention and $g_2 = \sqrt{x+1}$ for dynamic diminishing attention \footnote{The derivative of $\sqrt{x+1}$ becomes larger than that of $\log_a(x+1)$ when $x > 3 \log a$, {which rarely happens in summarization.}}. For a fair comparison, we train a PG model without their coverage mechanism, and apply our proposed diminishing attentions ({DimAttn} or {DyDimAttn}) from 230k iterations onward and train for 10k iterations for DimAttn and 15k iterations for DyDimAttn. We use the Adagrad optimizer with a learning rate of 0.15 for training PG and the same learning rate of 0.15 for training the coverage mechanism with a batch size of 32. We found that increasing the beam size from 4 to 6 leads to significant improvements to our model while it does not give any improvement to the baselines. We use a length normalization factor of $1.5$ for our method and apply trigram-blocking \citep{Paulus2018ICLR} during inference. 

\paragraph{BERT-based models}  The BERT-based models\citep{Liu2019TextSW} are trained with an initial learning rate of 2e-3 for encoder and 0.2 for decoder using the Adam optimizer. We fine-tune the models of \citet{Liu2019TextSW} for 10k updates for diminishing attention and 15k updates for dynamic diminishing attention using the same hyper-parameters as \citep{liu-lapata-2019-hierarchical}. We use $g=g_1=2.2$ and same as $g_2$ as PG-based models. We use a beam size of 5 and length penalty of 1. 

\paragraph{BART-based models} We finetune the BART-large model \citep{lewis2019bart}  with 406M parameters for 5 epochs using Adam optimizer with a similar learning rate schedule as \citet{lewis2019bart} - a learning rate of 3e-5 and 500 warm-up steps. Maximum token per GPU is set to 1024 and we accummulate gradients for 32 times for one update. 
We use a length normalization factor of 0.0 and a beam size of 4 and apply trigram-blocking for inference. We selected the exponent of $g$ in a range of $\{0.55, 0.65, 0.75\}$. We use the same exponent for $g_1$ as $g$ and select $g2$ in a range of $\{0.5, 0.6, 0.7\}$. We use $g = g_1 = (x+1)^{0.65}$ for diminishing attention and $g_2 = (x+1)^{0.6}$ for dynamic diminishing attention. 

\begin{table}[t!]
    \centering
     \caption{Summarization dataset statistics.}
    {\begin{tabular}{ c c c c} 
         \hline
         \textbf{Dataset} & CNN/DM & NYT \\
         \hline
         \textbf{Size} & 312,085 & 104,286  \\
         \textbf{Train/val/test} & 287,227/13,368/11,490 & 96,834/4,000/3,452 \\
         \hline
    \end{tabular}
    }
    \label{tab:dataset_stats}
\end{table}

\subsection{Machine translation}
The WMT14' train sets contain about 4.5 million instances for the En-De task and 35 million instances for the En-Fr task. We use newstest2013 as the validation set, and newstest2014 as the testing set. 
We use Adam optimizer with a learning rate of 0.0005 to fine-tune the baseline models for 300 updates for En-De and 3500 updates for En-Fr with 16 times of gradient accumulation for each update. 
On En-De, we use a beam size of 4 and length penalty factor of 0.7 for our diminishing attentions. On En-Fr, we use a beam size of 4 and length normalization factor of 0.6. We use the same $g$, $g_1$ and $g_2$ as the BART-based models. 

\subsection{Image-paragraph generation}
We use the standard split of 14,575/2,487/2,489(train/val/test) for the Stanford Image-Paragraph dataset \citep{hier-dataset}. We use Adam optimizer with a learning rate of 5e-4 and batch size of 20 for fine-tuning the cross-entropy baseline for 20 epochs and a learning rate of 6.7e-06 and batch size of 30 for fine-tuning the self-critical training baseline for 40 epochs. We decay the learning rate every 5 epochs and use a learning rate decay rate of 0.85. We select the log base of $g$ in a range of $\{1.9, 2.1, 2.3, 2.7\}$ and use the same log base for $g_1$ as $g$, and select $g_2$ in a range of $\{1.75, 1.95, 2.15, 2.35\}$. We use $g=\log_{1.9}(x+1)$ for diminishing attention and $g_1 = \log_{1.9}(x+1)$ and {$g_2 = \log_{1.95}(x+1)$} for dynamic diminishing attention for both cross-entropy and self-critical training. We follow the baselines in all the other hyperparameter settings.  

\section{Analysis}
\subsection{Ablation study on PG-based CNN/DM models}
\label{append:400-ablation}
We conduct an ablation study to analyze the impact of each component in our model.
Table \ref{table:ablation} shows the improvements for diminishing attentions and other components as they are added one at a time. By adding {DimAttn} or {DyDimAttn} only, our model outperforms the baselines. Length normalization and trigram-blocking further improve the ROUGE scores. 
\begin{table}[t!]
\setlength{\tabcolsep}{8pt}
\centering
\caption{Ablation study. All the models have the same settings as the PG baseline, \ie  the maximum encoding steps are all set to 400 (thus the difference in ROUGE scores from the {first} block in Table 1 in the main paper.}
{\begin{tabular}{lccc} 
\textbf{Model} &\bf{R-1} &\bf{R-2} &\bf{R-L}\\
\midrule
PG + Cov.           & 39.08 & 17.09 & 35.92 \\
PG + DimAttn             & 39.30 & 17.48 & 36.31 \\
~+ Length Norm.     & 39.64 & 17.51 & 36.61\\
~+ Trigram-Blocking & 39.92 & 17.64 & 36.88\\
\midrule
PG + Cov.           & 39.08 & 17.09 & 35.92 \\
PG + DyDimAttn           & 39.70 & 17.80 & 36.67\\
~+ Length Norm.     & 39.92 & 17.80 & 36.89\\
~+ Trigram-Blocking & 40.13 & 17.87 & 37.09\\

\bottomrule
\end{tabular}
}
\vspace{1em}
\label{table:ablation}
\end{table}

\subsection{N-gram repetition ratio of PG-based CNN/DM models}
\label{append:reptition}
\begin{figure}[t!]
    \centering
    \caption{\textbf{N-gram repetition} ratio of the model outputs for the PG baselines, diminishing attentions and reference summaries.}
    \hspace{-1em}\includegraphics[width=0.7\textwidth]{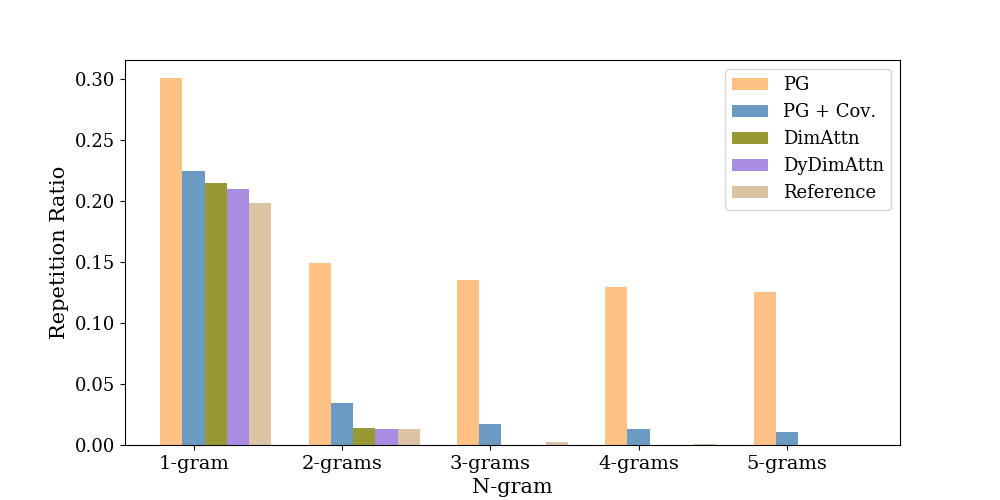}
    
    \label{fig:rep_ratio}
\end{figure}

We measure the \textit{repetition ratio} by calculating the duplicate n-grams in a summary. Figure \ref{fig:rep_ratio} shows the repetition ratio of summaries generated by PG baselines, our model and gold summaries. Our method yields significantly less repetition in terms of unigrams and bigrams compared to the vanilla PG. It outperforms the PG+Cov. model as well. trigram repetition is completely eliminated as trigram-blocking is applied. 

\subsection{Examples of summaries}
\label{append:examples}
We show two examples of summaries generated by the PG + Cov. model with trigram-blocking and our PG + DyDimAttn model in Table \ref{tab:example1} and Table \ref{tab:example2}. 
We also show two examples of summaries generated by BARTSum and BARTSum + DyDim model in 
Table~\ref{tab:bart-example1} and Table~\ref{tab:bart-example2}. 

\begin{table*}
\centering
\vspace*{10pt}
\small
\captionof{table}{The first example from the CNN/DM test set showing the outputs of PG + Cov. + Trigram-blocking and our model.}
\begin{tabular}{|p{\textwidth}|}
\toprule
{\bf Article}\\
\midrule
Ultimately , Bristol City were never destined to become the first football league club to win promotion this season at Deepdale, even though they are inching ever closer. Swindon's win over Peterborough meant Steve Cotterill will have to wait until Tuesday to finish the job of returning to the championship, but almost as importantly they made sure Preston wouldn't make any ground on them in the top two. Three points at Bradford on Tuesday will do the trick - only that standing in their way now.
\textit{(...)}\\
\midrule
{\bf Reference }\\
\midrule
Second-placed preston hosted league one leaders Bristol City. Jermaine Beckford fired the home side into the lead in the 59th minute. Aaron Wilbraham equalised for Bristol City four minutes later.
 \\
\midrule
{\bf PG + Cov. + Trigram-blocking }\\
\midrule
\hl{Swindon's win over Peterborough meant Steve Cotterill will have to wait until Tuesday to finish the job of returning to the championship, but almost as importantly they made sure Preston wouldn't make any ground on them in the top two.}
\hl{City had dispensed of their rivals on Tuesday night, hammering them 3-0, and took great delight} \hl{in their triumph.}
 \\
\midrule
{\bf PG + DyDimAttn} \\
\midrule
Bristol City beat Preston 3-0 in the premier league \hl{on Tuesday night.}
Aaron Wilbraham opened the scoring for the hosts in the second half.
The result means \hl{Steve Cotterill will have to wait until Tuesday to finish the job.}

\\
\bottomrule
\end{tabular}

\label{tab:example1}

\end{table*}

\begin{table*}
\vspace*{10pt}
\small
\captionof{table}{The second example from the CNN/DM test set showing the outputs of PG + Cov. + Trigram-blocking and our model. Highlighted spans are the phrases whose lengths are equal to or longer than 3 tokens and are copied verbatim from the source document.}
\begin{tabular}{|p{\textwidth}|}
\toprule
{\bf Article}\\
\midrule
\textit{(...)} Those years will always hang heavy around Wenger's neck. Yes, yes, there's the FA cup currently sitting there in the trophy cabinet alongside the shiny shield they won in august. And yes, yes, they are clear favourites to win it again. An FA cup isn't the definition of success for the great clubs of europe. Only the Wenger apologists can't see that. Olivier Giroud celebrates Arsenal's third goal by Alexis Sanchez in their win over Liverpool on Saturday. Arsene Wenger's gunners now have 15 wins from 17 games since new year's day, but it's come too late. So are gooners seduced by this end of season run they're currently on? It's impressive, no question. 15 wins in 17 games since their loss new year's day. But this is classic Arsene Wenger isn't it? After that defeat at Southampton on January 1 the gunners were sixth in the table , 13 points behind leaders Chelsea. 
\textit{(...)}\\
\midrule
{\bf Reference }\\
\midrule
Arsenal have a chance at the premier league title if Chelsea throw it away. The gunners are on a terrific run of \hl{15 wins in 17} since January 2 's loss. The two defeats in this spell were \hl{at Tottenham and at home to Monaco.} These \hl{were the two key games all arsenal fans desperately wanted to win.} Read : Arsenal have exactly the same record in league as last season. Click here for all the latest Arsenal news.
 \\
\midrule
{\bf PG + Cov. + Trigram-blocking }\\
\midrule
\hl{Arsene Wenger's gunners now have 15 wins from 17 games since new year's day, but it's come too late.}
\hl{Arsene Wenger isn't it ? After that defeat at Southampton on january 1 the gunners were sixth in the table, 13 points behind leaders chelsea.}
\hl{The FA cup currently sitting there in the trophy cabinet alongside the shiny shield they won in august.}
 \\
\midrule
{\bf PG + DyDimAttn} \\
\midrule
Arsenal \hl{were sixth in the table, 13 points behind leaders chelsea.}
Arsene Wenger's side \hl{have 15 wins from 17 games since new year's day.}
But this is not \hl{the definition of success for the great clubs of europe.}
Click here for Manchester United transfer news.
\\
\bottomrule
\end{tabular}
\label{tab:example2}
\end{table*}

\begin{table*}
\centering
\vspace*{10pt}
\small
\captionof{table}{The first example from the CNN/DM test set showing the outputs of the BARTSum and BARTSum + DyDim.}
\begin{tabular}{|p{\textwidth}|}
\toprule
{\bf Article}\\
\midrule
The NSW State Emergency Service (SES) are warning the public that scammers are making calls falsely claiming to fundraise, abusing people’s goodwill as NSW is ravaged by wild weather. The public have been alerted to the heartless scheme via an important notice issued on the NSW SES Facebook page. ‘The NSW SES does NOT request donations from the public,’ NSW SES assure the public. ‘We will never call and ask for money. ‘If you are receiving calls of this nature, please be advised it is a scam.’ The public have been alerted to the heartless scheme via an important notice issued on the NSW SES Facebook page. NSW SES have received more than 6500 requests for help since the storms began on Monday, with flash flooding, trees down and power outages across the Sydney, Newcastle and Hunter regions.\textit{(...)} People have reacted with disgust to the attempt to con people out of their money, condemning those responsible for trying to benefit from a crisis. 
\textit{(...)}\\
\midrule
{\bf Reference }\\
\midrule
NSW SES warns scammers are phoning people claiming to fundraise. The State Emergency Service say they never call and ask for money. People have responded with disgust at the heartless. Con artists are taking advantage of people's goodwill as NSW is facing severe weather conditions. NSW SES have received more than 6500 requests for help since Monday.
 \\
\midrule
{\bf BARTSum }\\
\midrule
The NSW State Emergency Service (SES) are warning the public that scammers are making calls falsely claiming to fundraise. The public have been alerted to the heartless scheme via an important notice issued on the NSW SES Facebook page. People have reacted with disgust to the attempt to con people out of their money, condemning those responsible for trying to benefit from a crisis. ‘Oh my god! Who are these absolute mongrels that take advantage of these sorts of situations!’ wrote Belinda Weston.
 \\
\midrule
{\bf BARTSum + DyDimAttn} \\
\midrule
The NSW State Emergency Service (SES) are warning the public that scammers are making calls falsely claiming to fundraise, abusing people’s goodwill as NSW is ravaged by wild weather. The public have been alerted to the heartless scheme via an important notice issued on the NSW SES Facebook page. People have reacted with disgust to the attempt to con people out of their money. More than 6500 requests for help since the storms began on Monday, with flash flooding, trees down and power outages across the Sydney, Newcastle and Hunter regions.
\\
\bottomrule
\end{tabular}
\label{tab:bart-example1}

\end{table*}
\begin{table*}
\centering
\vspace*{10pt}
\small
\captionof{table}{The second example from the CNN/DM test set showing the outputs of the BARTSum and BARTSum + DyDim.}
\begin{tabular}{|p{\textwidth}|}
\toprule
{\bf Article}\\
\midrule
(CNN)"Jake the dog and Finn the human. The fun will never end. Adventure Time." So begins the dreamy theme song intro to the strangely addictive Cartoon Network TV show that's centered around psychedelic characters like the Ice King, Marceline the Vampire Queen and, of course, Jake and Finn. Now, mega-fans of the hit show can experience "Adventure Time" in the skies. Thai Smile, a subsidiary of Thailand flag carrier Thai Airways, on Thursday unveiled colorful new livery featuring Jake, Finn and the beloved Princess Bubblegum sprawled across an Airbus A320 at Bangkok's Suvarnabhumi International Airport. The interior of the plane also has an Adventure Time theme, with overhead bins, head rests and even air sickness bags covered in the faces of characters from the show. Airlines show off their new flying colors. The Adventure Time plane is the result of a partnership between Thai Airways subsidiary Thai Smile and Cartoon Network Amazone, a new water park near the Thai resort city of Pattaya featuring attractions based on shows that appear on the Turner Broadcasting System channel. Turner Broadcasting is a parent company of CNN. Check out these cool airline liveries. The inaugural Thai Smile Adventure Time flight takes place on April 4, heading from Bangkok to Phuket.
\textit{(...)}\\
\midrule
{\bf Reference }\\
\midrule
Thai Airways subsidiary Thai Smile features Cartoon Network paint job on A320 jet. Overhead bins, head rests and air sick bags feature characters from Cartoon Network.
 \\
\midrule
{\bf BARTSum }\\
\midrule
Thai Smile unveils colorful new livery featuring Jake, Finn and the beloved Princess Bubblegum. The interior of the plane also has an Adventure Time theme, with overhead bins, head rests and even air sickness bags covered in the faces of characters. The inaugural Thai Smile Adventure Time flight takes place on April 4, heading from Bangkok to Phuket.
 \\
\midrule
{\bf BARTSum + DyDimAttn} \\
\midrule
Thai Airways subsidiary Thai Smile unveils "Adventure Time" livery for new Airbus A320. The interior of the plane also has an Adventure Time theme. The plane is the result of a partnership between Thai Smile and Cartoon Network Amazone. The inaugural Thai Smile Adventure Time flight takes place on April 4.
\\
\bottomrule
\end{tabular}

\label{tab:bart-example2}

\end{table*}


\subsection{Abstractiveness ratio}
\label{append:abstractiveness_ratio}
In Table~\ref{table:novelcnn}, we present the  \textit{abstractiveness} of the summaries generated by PG-based models. The baseline model equipped with dynamic diminishing attention\footnote{We omit the abstractiveness of PG + DimAttn as there is a less significant effect.} has a higher abstractiveness against the baseline.  

We present the abstractiveness of the summaries generated by the BERT-based models on NYT in Table~\ref{table:novelnyt}. 

The same consecutive tokens in the source document tend to be generated by the summarization models. This could be because that the source sequence is perfectly fluent and is favored by the language model. 
However, the diminishing attentions imposed by the submodular coverage inform the model to have more appropriate generation for better coverage, which is not necessarily exactly the same as the source sequence. 

\begin{table}[h]
\setlength{\tabcolsep}{8pt}


\end{table} 

\begin{figure}[t!]
\begin{minipage}[t]{0.48\textwidth}
\centering 
\captionof{table}{Novel n-gram percentage on CNN/DM (PG-based models).}
\label{table:novelcnn}
\resizebox{0.98\columnwidth}{!}{\begin{tabular}{lccccc} 
\textbf{Model} &\bf{1-gr} &\bf{2-gr} &\bf{3-gr} & \bf{4-gr} & \bf{5-gr} \\
\midrule
Reference         & 13.6 & 49.0 & 67.7 & 76.9 & 82.3 \\
\midrule
PG + Cov.         & \bf{0.3} & 4.0 & 7.1 & 9.7 & 12.2 \\
PG + DyDimAttn    & \bf{0.3} & \bf{4.3} & \bf{7.5} & \bf{10.2} & \bf{12.8} \\
\bottomrule
\end{tabular}
}
\end{minipage}
\hfill
\begin{minipage}[t]{0.48\textwidth}
\centering 
\captionof{table}{Novel n-gram percentage on NYT (BERT-based models).}
\label{table:novelnyt}
\resizebox{1\columnwidth}{!}{\begin{tabular}{lccccc} 
\textbf{Model} &\bf{1-gr} &\bf{2-gr} &\bf{3-gr} & \bf{4-gr} & \bf{5-gr} \\
\midrule
Reference         & 21.7 & 51.9 & 66.8 & 75.4 & 81.3 \\
\midrule
BertSum         & 4.9 & 22.3 & 36.4 & 47.3 & 56.0 \\
BertSum + DyDimAttn      & \bf{5.0} & \bf{22.6} & \bf{36.9} & \bf{47.8} & \bf{56.6} \\
\bottomrule
\end{tabular}
}
\end{minipage}
\end{figure}

\subsection{Human evaluation details}

\label{append:human_eval_interface} 
\begin{figure}[ht]
\vspace{-0.5em}
    \centering
    \caption{Human evaluation interface.}
    \includegraphics[width=1\textwidth]{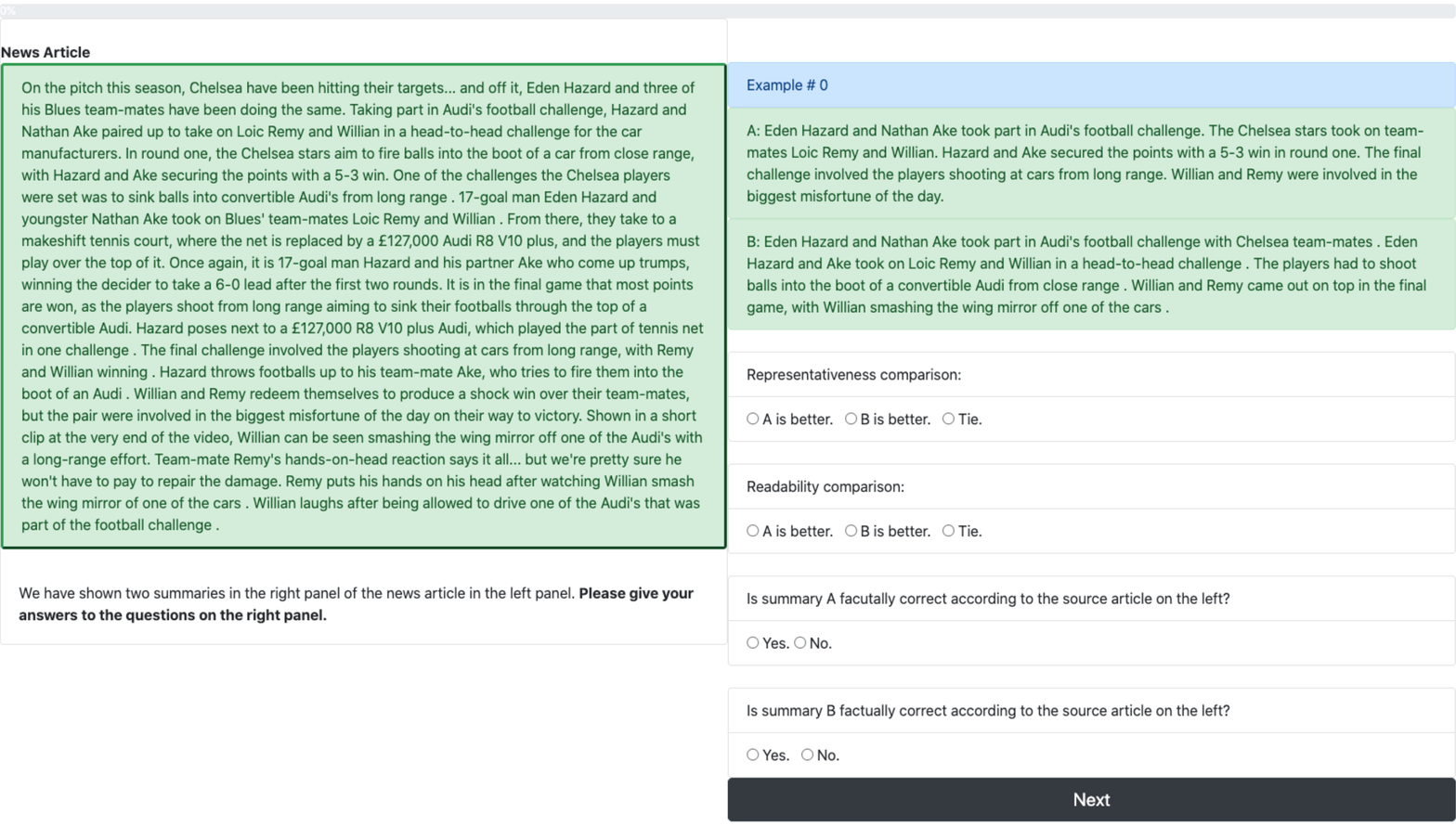}
    \vspace{-2em}
    \label{fig:human_eval_web}
\end{figure}

We show the interface used for human evaluation in Figure \ref{fig:human_eval_web}. 

